\documentclass[letterpaper, 10 pt, conference]{ieeeconf}  

\IEEEoverridecommandlockouts                              

\usepackage{url} 
\usepackage[hidelinks]{hyperref}
\usepackage{breakurl}
\usepackage{graphicx} 
\usepackage{multirow} 
\usepackage{multicol} 
\usepackage{gensymb} 
\usepackage{amssymb}
\usepackage{amsmath}
\usepackage{xcolor}
\usepackage{siunitx}
\usepackage{algorithm}
\usepackage{algpseudocode}
\usepackage{booktabs}
\usepackage{pgffor}
\usepackage{pgfmath}
\usepackage{xparse}
\usepackage{etoolbox}
\usepackage{tikz}
\usetikzlibrary{positioning}
\usetikzlibrary{calc}
\tikzset{>=latex}
\usepackage[nolist,nohyperlinks]{acronym}
\begin{acronym}[AAAAAAAAA]
    \acro{cad}[CAD]{Computer-Aided Design}
    \acro{cnn}[CNN]{Convolutional Neural Network}
    \acro{dof}[DoF]{Degree of Freedom}
    \acro{hdr}[HDR]{High-Dynamic Range}
    \acro{leo}[LEO]{Low Earth Orbit}
    \acro{port}[LM-MAP]{Lockheed Martin Mission Augmentation Port}
    \acro{ransac}[RANSAC]{RANndom SAmple Consensus}
    \acro{slam}[SLAM]{Simultaneous Localisation And Mapping}
    \acro{vo}[VO]{Visual Odometry}
\end{acronym}

\def\subfigstyle{\footnotesize}
\title{\LARGE \bf Mixing Data-driven and Geometric Models for Satellite Docking Port State Estimation using an RGB or Event Camera}

\author{Cedric Le Gentil$^{1}$, Jack Naylor$^{2}$, Nuwan Munasinghe$^{1}$, Jasprabhjit Mehami$^{2}$,\\Benny Dai$^{1}$, Mikhail Asavkin$^{3}$, Donald G. Dansereau$^{2}$, Teresa Vidal-Calleja$^{1}$
\thanks{*This work was supported by NSW Space Research Network, grant ID RP220201}
\thanks{$^{1}$Robotics Institute,
        University of Technology Sydney, Australia.}%
\thanks{$^{2}$Australian Centre for Robotics, School of Aerospace, Mechanical and Mechatronic Engineering, University of Sydney, Australia.}%
\thanks{$^{3}$ANT61, Australia.}
\thanks{Corresponding author: {\tt\small cedric.legentil@uts.edu.au}}%
}

\begin{document}

\maketitle
\thispagestyle{empty}
\pagestyle{empty}

\begin{abstract}
In-orbit automated servicing is a promising path towards lowering the cost of satellite operations and reducing the amount of orbital debris.
For this purpose, we present a pipeline for automated satellite docking port detection and state estimation using monocular vision data from standard RGB sensing or an event camera.
Rather than taking snapshots of the environment, an event camera has independent pixels that asynchronously respond to light changes, offering advantages such as high dynamic range, low power consumption and latency, etc.
This work focuses on satellite-agnostic operations (only a geometric knowledge of the actual port is required) using the recently released Lockheed Martin Mission Augmentation Port (LM-MAP) as the target.
By leveraging shallow data-driven techniques to preprocess the incoming data to highlight the LM-MAP's reflective navigational aids and then using basic geometric models for state estimation, we present a lightweight and data-efficient pipeline that can be used independently with either RGB or event cameras.
We demonstrate the soundness of the pipeline and perform a quantitative comparison of the two modalities based on data collected with a photometrically accurate test bench that includes a robotic arm to simulate the target satellite's uncontrolled motion. 

\end{abstract}

\section{Introduction}

\begin{figure}
    \centering
    \def\imgscale{0.42}
    \def\imgwidth{\imgscale\columnwidth}
    \def\imgdist{0.05\columnwidth}
    \def\subfigdist{0.05cm}
    \def\legendstyle{\scriptsize \color{red}}
    \def\vdist{0.6cm}
    \begin{tikzpicture}
        \tikzstyle{img} = [fill=white, rectangle, align = center, execute at begin node=\setlength{\baselineskip}{8pt}, inner sep=0, outer sep=0]
        \tikzstyle{subfig} = [fill=white, rectangle, align = center, text width = \imgwidth,  minimum width = \imgwidth, execute at begin node=\setlength{\baselineskip}{8pt}, inner sep=0, outer sep=0]
        \tikzstyle{legend} = [rectangle, align = center, text width = 5em, execute at begin node=\setlength{\baselineskip}{8pt}, inner sep=0, outer sep=0, anchor=north]

        \node[img](diagram){\includegraphics[width=\imgwidth,clip]{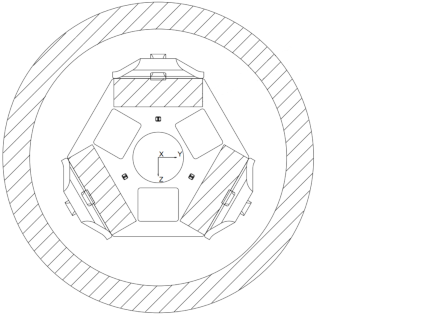}};
        \node[subfig, below=\subfigdist of diagram](subdiagram){\subfigstyle(a) LM-MAP diagram};

        \node[img, below=\vdist of diagram](rgb){\includegraphics[width=\imgwidth,clip]{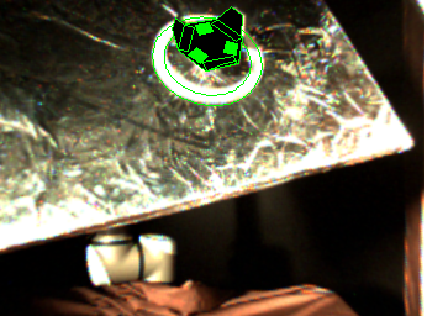}};
        \node[subfig, below=\subfigdist of rgb]{\subfigstyle(c) RGB-based detection};
        
        \node[img, right=\imgdist of rgb](event){\includegraphics[width=\imgwidth,clip]{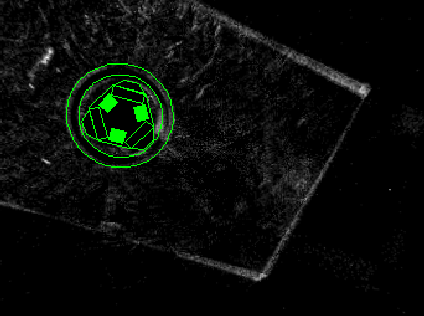}};
        \node[subfig, below=\subfigdist of event]{\subfigstyle(d) Event-based detection};
        
        \node[img, right=\imgdist of diagram](photo){\includegraphics[width=\imgwidth,clip]{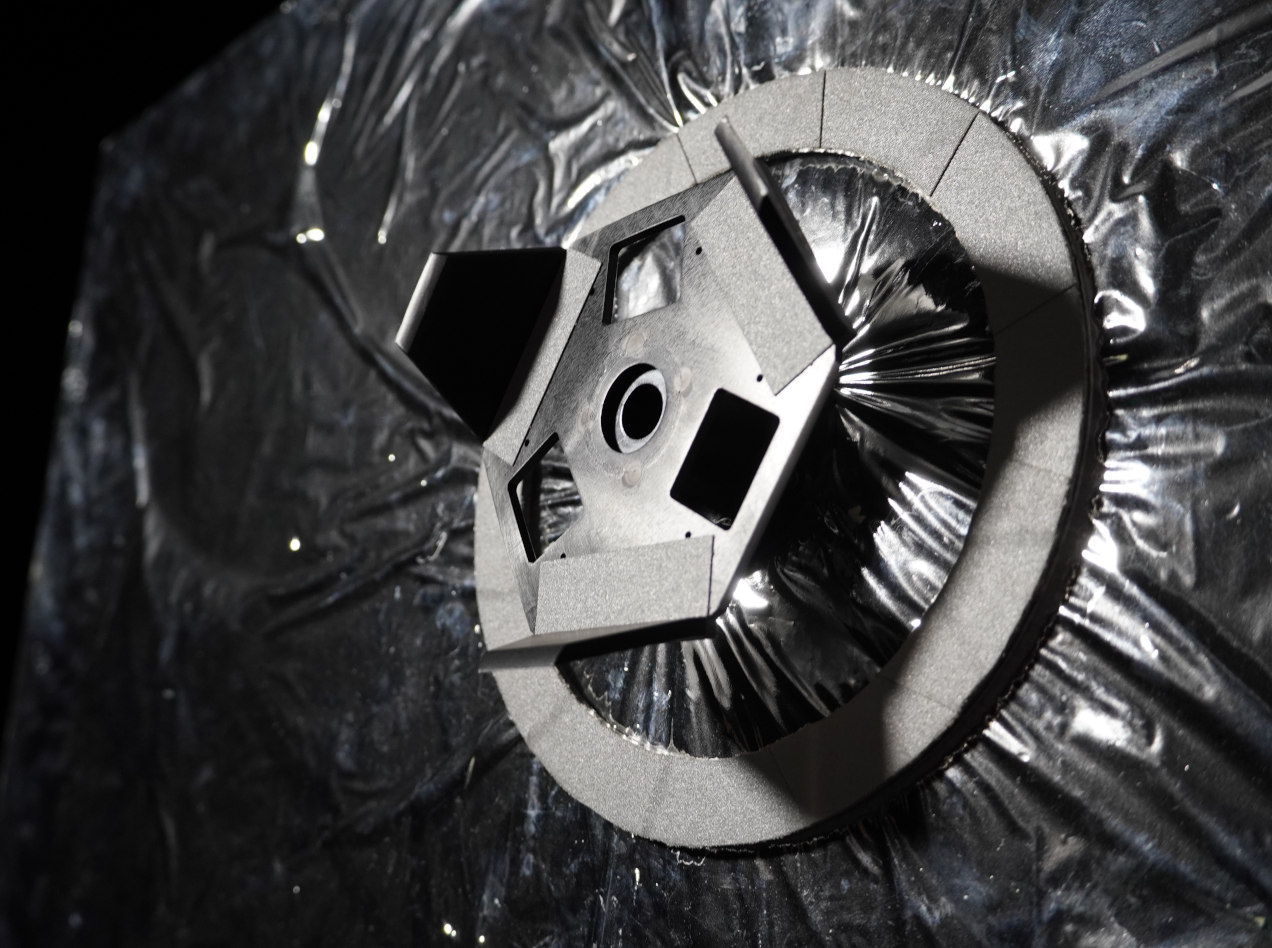}};
        \node[subfig, right=\imgdist of subdiagram]{\subfigstyle(b) LM-MAP mock-up};

        \node[legend, below=0.05cm of diagram.north east, xshift=-0.6cm](ringlegend){\legendstyle Reflective ring\\navigational aid};
        \node[legend, above=0.05cm of diagram.south east, xshift=-0.6cm](reflectorlegend){\legendstyle Port's reflective\\markers};

        \draw[->, red] (ringlegend.south) -- (\imgscale*1.6, \imgscale*0.8);
        \draw[->, red] (reflectorlegend.north west) -- (\imgscale*-2.2, \imgscale*-0.8);
        \draw[->, red] (reflectorlegend.north west) -- (\imgscale*-0.3, \imgscale*-0.8);
        \draw[->, red] (reflectorlegend.north west) -- (\imgscale*-0.9, \imgscale*1.2);
        
    \end{tikzpicture}
    \vspace{-0.25cm}
    \caption{The proposed method performs the detection and state estimation of the Lockheed Martin Mission Augmentation Port (LM-MAP) ((a) and (b)) using standard RGB images (c) or event-based data (d).}
    \label{figure:teaser}
\end{figure}
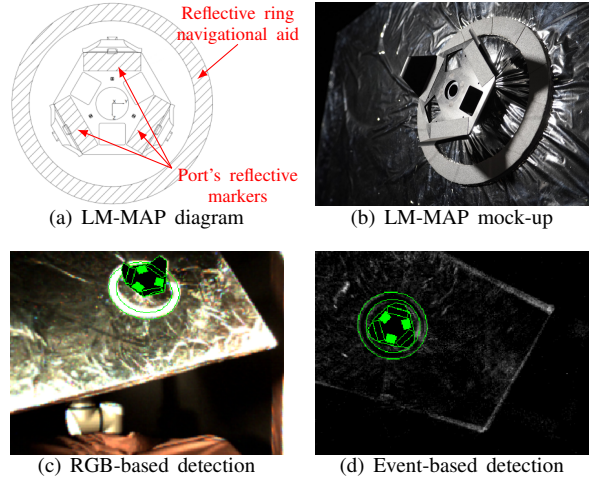

Satellite operations support a wide range of infrastructure essential to today's society.
First used as scientific, technological, and military demonstrators during the Cold War, satellites are now part of our daily life (e.g., telecommunication, GPS, etc) and are shown to be valuable tools for environment monitoring especially in the effort to fight climate change \cite{yang2013climate}.
Unfortunately, the growing number of satellites in orbit comes with logistic and obsolescence problems given the accumulation of debris and decommissioned satellites.
Along with the financial incentive to reduce the number of rocket launches, there is a growing interest for in-orbit maintenance of satellites to extend their lifetime and avoid cluttering orbits. 
In 2022, Lockheed Martin released the specification of a docking port \cite{LockheedMartin2022LockheedStandard} to standardise and facilitate in-orbit servicing.
In this paper, we work towards the adoption of such a standard by addressing the issue of the \ac{port} detection and state estimation for autonomous docking operation as illustrated in Fig.~\ref{figure:teaser}.

While some systems are tested in space \cite{Samson2004NeptecApplications, Ruel2010On-orbitSensor}, algorithmic breakthroughs for autonomous docking or in-orbit rendezvous heavily rely on the development of software \cite{ma2006validation} and physical \cite{zebenay2015hardware,dlr2017epos} simulators here on Earth.
In \cite{Oumer2015Vision-basedSatellite} the authors focus on camera-based docking port detection and localisation with a large-scale (20m) physical simulator that consists in two industrial robotic arms and a rail system.
In previous work \cite{munasinghe2024towards}, we presented our photometrically accurate real-world satellite-docking simulator to allow for the design of novel vision-based perception algorithms.
This test bench allows us to compare the performance of the proposed algorithm with both RGB and event cameras.

Event cameras, also called neuromorphic cameras \cite{Lichtsteiner2008DVS}, introduced a novel way to acquire ``visual" data.
Unlike traditional cameras where all the pixels are triggered simultaneously to get a snapshot of the environment, the pixels of an event camera independently trigger events when the level of light changes in that pixel.
Accordingly, the output of an event camera consists of an asynchronous stream of events (timestamp, x and y pixel location, and direction of change) that display interesting properties such as low latency, \ac{hdr}, absence of motion blur, etc.
Naturally, the robotics community has developed an increasing number of algorithms over the past decade to leverage event vision in various applications \cite{Gallego2022Event-BasedSurvey}, with examples for keypoint detection~\cite{alzugaray18}, tracking~\cite{Zhu2017, alzugaray2020, legentil2023}, odometry \cite{rebecq2017evo,legentil2020idol}, and SLAM \cite{Vidal2018UltimateScenarios}.


While the state of event-based research is not as mature as the standard camera counterpart, there has already been a push toward space-oriented uses of event vision.
For example, in \cite{Chin2019StarCamera}, \cite{Bagchi2020Event-basedTransforms}, \cite{Cohen2019Event-basedAwareness}, and \cite{Afshar2020Event-BasedAwareness} the focus is the detection and tracking of point object/celestial bodies.
Closer to the proposed work, \cite{Jawaid2023TowardsSensing} proposes an approach to perform satellite state estimation using neural networks to extract keypoints that can later be associated with their corresponding vertices in the \ac{cad} model.
The focus in \cite{Jawaid2023TowardsSensing} is closing the domain gap between simulated and real data.
Reducing the ``sim-to-real" gap allows for network training with large amount of data without the caveat of collecting real-world data.

Outside of event-only spacecraft detection and localisation, numerous works are based on deep learning as discussed in a recent survey \cite{pauly2023deeplearningsurvay}.
Such methods are specifically trained for particular models of satellite like in \cite{Jawaid2023TowardsSensing}, where the satellite's \ac{cad} model is required during training.
Thus, they do not generalise to new targets and require retraining.
Also, this might not be compatible with certain servicing or deorbiting missions due to alterations or damages to the satellite during its years of service (potential difference between the \ac{cad} model and the real satellite).
Most deep learning approaches in \cite{pauly2023deeplearningsurvay} rely on network models with millions of parameters possibly making it difficult to deploy for real-time computation on embedded hardware.
Another interesting approach is the online supervision in \cite{park2024onlinesupervised} with an adaptive Kalman filter to learn the network parameters while performing the satellite approach manoeuvres.
However, this method also requires a prior \ac{cad} model of the spacecraft.

In this paper, we propose a \ac{port} detection and monocular state estimation framework that can handle either standard camera images or event data.
The core principle of our approach is the combination of data-driven and basic geometric models for data-efficient lightweight estimation.
The data-driven component highlights key features of the \ac{port} that are the navigation ring and the port reflectors as seen in Fig.~\ref{figure:teaser}.
Then, simple geometric models with a \ac{ransac} approach allow for 6-\ac{dof} estimation of the port's pose.
The contributions of this work are the design and implementation of a satellite-agnostic docking port detection and localisation algorithm, the evaluation of the proposed framework with both event and RGB data, and the open-source release of the datasets used in the evaluation.

\section{Method}

\subsection{Overview}

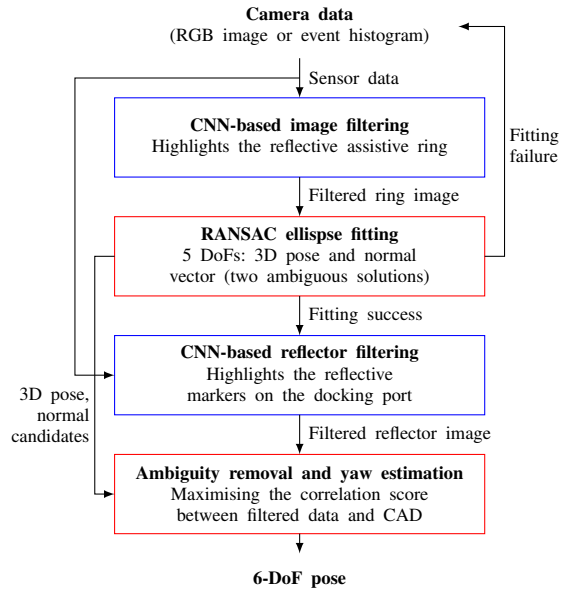
\begin{figure}
    \centering
    \def\hdist{5em}
\def\vdistlong{2.5em}
\def\vdist{1.5em}
\def\blockheight{3.0em}
\def\blockwidth{14.0em}
\def\innerpad{0.0em}
\def\textsize{\scriptsize}
\def\arrowtextsize{\scriptsize}
\begin{tikzpicture}[auto]
    \tikzstyle{input} = [fill=white, rectangle, minimum height = 2.4em, text width = \blockwidth-2em,  minimum width = \blockwidth-2em, align = center, node distance = 5em, execute at begin node=\setlength{\baselineskip}{8pt}, inner sep=\innerpad, outer sep=0]
    \tikzstyle{block} = [draw, fill=white, rectangle, minimum height = \blockheight, text width = \blockwidth,  minimum width = \blockwidth, align = center, inner sep=\innerpad, outer sep=0, node distance = 11em, execute at begin node=\setlength{\baselineskip}{8pt}] 
    \tikzstyle{learntblock} = [block, draw=blue]
    \tikzstyle{modelblock} = [block, draw=red]
    \tikzstyle{arrowtext} = [rectangle, text width = (0.5*\blockwidth),  minimum width = (0.5*\blockwidth), align = left, execute at begin node=\setlength{\baselineskip}{8pt}] 

    \node[input] (cam) {\textsize \textbf{Camera data}\\(RGB image or event histogram)};

    \node[learntblock, below=\vdist of cam] (ring) {\textsize \textbf{CNN-based image filtering}\\Highlights the reflective assistive ring};
    \node[modelblock, below=\vdist of ring] (ellipse) {\textsize \textbf{RANSAC ellispse fitting}\\5 DoFs: 3D pose and normal vector (two ambiguous solutions)};
    \node[learntblock, below=\vdist of ellipse] (reflectors) {\textsize \textbf{CNN-based reflector filtering}\\Highlights the reflective markers on the docking port};
    \node[modelblock, below=\vdist of reflectors] (yaw) {\textsize \textbf{Ambiguity removal and yaw estimation}\\Maximising the correlation score between filtered data and CAD};

    \node[input, below=0.5*\vdist of yaw, minimum height=2.0em] (pose) {\textsize \textbf{6-DoF pose}};

    \draw[->] (cam) -- node[arrowtext]{\arrowtextsize Sensor data} (ring);
    \draw[->] (ring) -- node[arrowtext]{\arrowtextsize Filtered ring image} (ellipse);
    \draw[->] (ellipse) -- node[arrowtext]{\arrowtextsize Fitting success} (reflectors);
    \draw[->] (reflectors) -- node[arrowtext]{\arrowtextsize Filtered reflector image} (yaw);
    \draw[->] (yaw) -- node[arrowtext]{} (pose);

    \draw[->] ($(cam.south)!0.5!(ring.north)$) -| ($(reflectors.west)+(-\vdist,0)$) -- (reflectors.west);
    \draw[->] (ellipse.west) -| node[arrowtext, left, align=right, minimum width=3em, text width=3em, inner sep=2, outer sep=0, yshift=-(2*\vdist+\blockheight)]{\arrowtextsize 3D pose, normal candidates} ($(yaw.west)+(-0.5*\vdist,0)$) -- (yaw.west);
    
    \draw[->] (ellipse.east) -- ($(ellipse.east)+(0.5*\vdist,0)$) |- node[arrowtext, right, minimum width=3em, text width=3em, inner sep=2, outer sep=0, yshift=-(\vdist+\blockheight)]{\arrowtextsize Fitting failure} (cam.east);
\end{tikzpicture}
    \vspace{-0.5cm}
    \caption{Diagram overview of the proposed pipeline for satellite docking port detection and state estimation. The blue blocks are built upon data-driven techniques while the red block correspond to geometry-based algorithms.}
    \label{figure:overview}
\end{figure}

Let us consider a camera (RGB or event-based) and a \ac{port} moving freely in space.
The proposed pipeline aims at detecting the \ac{port} and estimating its 6-\ac{dof} pose in the camera reference frame.
As illustrated in Fig.~\ref{figure:overview}, it relies on a combination of data-driven and model-based techniques and can handle either event or RGB data sources independently: \acp{cnn} are used to filter the camera data by highlighting key components of the \ac{port} before fitting simple geometric models to perform pose estimation.

\begin{figure}
    \centering
    \def\cnnwidth{0.12\columnwidth}
    \def\imgwidth{0.3\columnwidth}
    \def\imgdist{0.01\columnwidth}
    \def\subfigdist{0.35cm}
    \def\legendstyle{\scriptsize \color{white}}
    \def\legenddist{-0.3cm}
    \def\vdist{0.5cm}
    \begin{tikzpicture}
        \tikzstyle{img} = [fill=white, rectangle, align = center, execute at begin node=\setlength{\baselineskip}{8pt}, inner sep=0, outer sep=0]
        \tikzstyle{subfig} = [fill=white, rectangle, align = center, text width = \columnwidth,  minimum width = \columnwidth, execute at begin node=\setlength{\baselineskip}{8pt}, inner sep=0, outer sep=0]
        \tikzstyle{legend} = [rectangle, align = left, text width = \imgwidth-0.2cm,  minimum width = \imgwidth-0.2cm, execute at begin node=\setlength{\baselineskip}{8pt}, inner sep=0, outer sep=0, anchor=north]
        \tikzstyle{cnntext} = [rectangle, align = center, text width = \cnnwidth,  minimum width = \cnnwidth, execute at begin node=\setlength{\baselineskip}{8pt}, inner sep=0, outer sep=0]

        \node[img] (event) {\includegraphics[width=\imgwidth,clip]{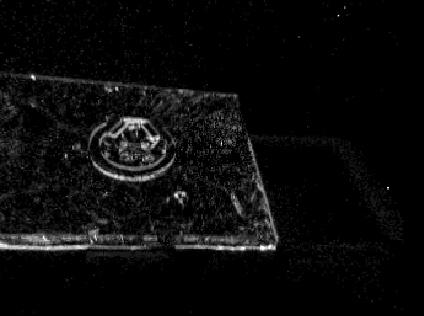}};
        \node[img, right=\imgdist of event] (eventcnn) {\includegraphics[width=\cnnwidth,clip]{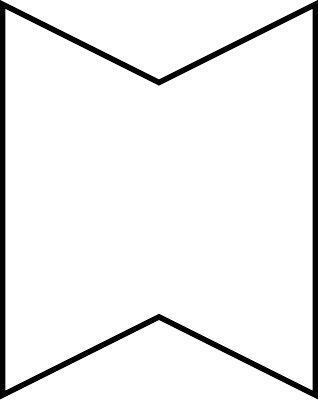}};
        \node[cnntext, right=\imgdist of event] {\textbf{CNN-event}};
        \node[img, right=\imgdist of eventcnn] (ringfiltered) {\includegraphics[width=\imgwidth,clip]{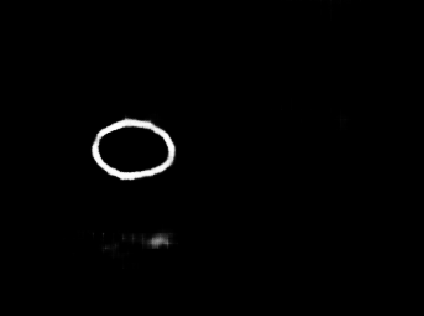}};

        \node[subfig, below=\subfigdist of eventcnn] {\subfigstyle (a) Ring filtering example with event-based input};

        \node[legend, above=\legenddist of event] {\legendstyle\textbf{Event histogram}};
        \node[legend, above=\legenddist of ringfiltered] {\legendstyle\textbf{Filtered image $\mathcal{I}_O$}};

        \node[img, below=\vdist of event] (rgb) {\includegraphics[width=\imgwidth,clip]{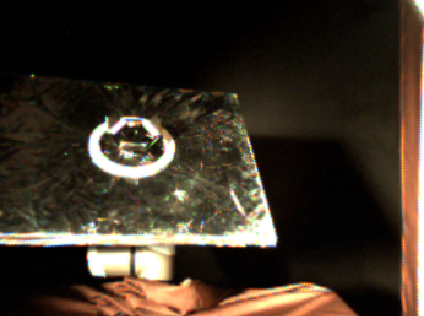}};
        \node[img, right=\imgdist of rgb] (rgbcnn) {\includegraphics[width=\cnnwidth,clip]{figures/cnn.pdf}};
        \node[cnntext, right=\imgdist of rgb] {\textbf{CNN-RGB}};
        \node[img, right=\imgdist of rgbcnn] (reflectorfiltered) {\includegraphics[width=\imgwidth,clip]{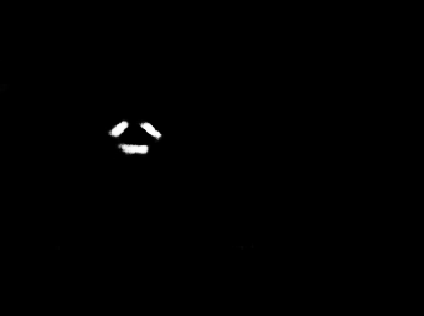}};

        \node[subfig, below=\subfigdist of rgbcnn]{\subfigstyle (b) Reflector filtering example with RGB input};
        
        \node[legend, above=\legenddist of rgb] {\legendstyle\textbf{Standard RGB image}};
        \node[legend, above=\legenddist of reflectorfiltered] {\legendstyle\textbf{Filtered image $\mathcal{I}_R$}};
    \end{tikzpicture}
    \vspace{-0.5cm}
    \caption{Illustration of the ring and reflector CNN-based filters. (a) shows a ring filtering example with event-based input ($N$ = 35k). (b) shows the reflector filtering with RGB data.}
    \label{figure:network_example}
\end{figure}

Concretely, images are collected with the camera either as RGB images or as the accumulation of $N$ events in an image-like histogram.
The sensor data is passed through a \emph{ring filter} \ac{cnn} that highlights the reflective assistive ring present around the \ac{port} as shown in Fig.~\ref{figure:network_example}(a).
By binarising and skeletonising the filtered image, an ellipse is fitted to the pre-processed sensor data.
Analysing the geometric characteristics of the ellipse given the actual size of the ring allows for the estimation of 5 \acp{dof} of the \ac{port} pose: the 3D position in the camera frame and the direction of the normal vector of the \ac{port}.
Note that the 2 \acp{dof} of the normal vector also present an ambiguity as two different vectors can explain the observed ellipse.
To solve for the ambiguity and estimate the last DoF of the \ac{port}'s pose, a second \ac{cnn} is used to highlight the three reflective rectangular markers present on the surface of the \ac{port}.
By computing a correlation score between the filtered image and the projection of a simplistic \ac{cad} model of the \ac{port}, the optimum ``yaw" angle around the normal vector is determined.

\subsection{CNN image filtering}

The proposed pipeline can be used either with a single RGB or event camera.
While RGB cameras directly provide images, the event data stream is not an image-like representation suitable for a standard \ac{cnn} input.
Accordingly, when using an event camera, the raw stream is converted into a succession of frames that are image-like histograms in which each bin corresponds to a pixel, and is populated with the events that occurred at this location (regardless of their polarity).
A total of $N$ consecutive events are used to generate the histograms ($N$=35k in our implementation).
Fig.~\ref{figure:network_example}(a) provides an example of such a histogram.

The filtering of the assistive navigational ring and the three reflective markers rely on a single \ac{cnn} architecture.
Similarly, the architecture is the same for RGB and event-based sensing except for the first layer which has three channels for RGB data as opposed to only one for event-based data.
As per the goal of performing satellite-agnostic \ac{port} detection, the proposed pipeline leverages a U-Net-like~\cite{Ronneberger2015U-Net:Segmentation} \ac{cnn} to infer mask-like images that highlight the appropriate \ac{port} features (navigation ring or reflectors) as illustrated in Fig.~\ref{figure:network_example}.
The chosen architecture consists of three consecutive convolution layers with ReLU activation functions and max-pooling, followed by three deconvolution layers with ReLU activation.
The one-channel output image has the same size as the input image and is passed through a sigmoid function to provide a mask with values between zero and one.
We denote $\mathcal{I}_O$ the output of the ring filter and $\mathcal{I}_R$ the one of the reflector filter.

The lack of fully connected layers implies that only local information is used in the filter output.
Accordingly, the proposed network cannot learn the ring position by ``recognising" a certain satellite's side panel and infer the \ac{port} position relative to the in-built knowledge of the specific satellite.
By only considering local information our approach is truly satellite-agnostic and avoids the risk of overfitting to a closed set of satellites.

\subsection{Ellipse fitting and 5-DoF estimation}

As shown in Fig.~\ref{figure:network_example}, the proposed pipeline leverages the reflective ring (navigational aid) that is present around the \ac{port} to estimate 5 of the 6 \acp{dof} of the \ac{port} pose.
Based on the standard pinhole camera model, the projection of a 3D circle on the image plane almost matches the shape of an ellipse.
Thus, we propose to use the simple conic section implicit representation 
\begin{equation}
    Ax^2 + Bxy + Cy^2 + Dx + Ey + F = 0
    \label{eq:conic}
\end{equation}
in 2D to approximate the ring projection in the image and subsequently perform the 5-\ac{dof} estimation of the \ac{port}'s state.

Concretely, given the ring-filtered image $\mathcal{I}_O$ we apply binarisation and skeletonisation~\cite{zhang1984skeleton} to obtain a one-pixel-wide representation of the scene that we denote $\mathcal{I}_S$.
The absence of a ring in the current view results in a very low number of active pixels in $\mathcal{I}_S$.
Accordingly, with a threshold $\gamma_S$, the estimation process is aborted if $\mathcal{I}_S<\gamma_S$.
Otherwise, the active pixels are converted into 2D points $\mathbf{X}\in\mathbb{R}^{2\times N}$, based on their image coordinates, and used in a \ac{ransac}-based ellipse fitting algorithm to estimate the parameters $\mathbf{e}_h = \begin{bmatrix}
    A&B&C&D&E
\end{bmatrix}$ in \eqref{eq:conic}.
Note that \eqref{eq:conic} possesses six parameters while an ellipse only has five \acp{dof}.
Thus, the value of $F$ is set as a constant equal to one.
The \ac{ransac} process
consists in selecting five points $\mathbf{X}_h \in \mathbb{R}^{2\times 5}$ and estimating the corresponding hypothetical ellipse parameters $\mathbf{e}_h$ by solving a linear problem.
From \eqref{eq:conic}, the conic representation can be converted into a minor/major axis ellipse representation.
After a sanity check on the ratio between the minor and major axis of the current hypothesis, the set of inliers from $\mathbf{X}$ is computed.
The \ac{ransac} process is repeated $H$ times or until the number of inliers reaches the number of active pixels in $\mathcal{I}_s$.
Eventually, the hypothesis associated with the most inliers is refined using all the inliers and selected for the rest of the estimation process.

Based on the extremities of the ellipse's major axis $\mathbf{x}_{M1}$ and $\mathbf{x}_{M2}$ in the image, the 3D position $\mathbf{p}$ of the \ac{port} is estimated as the center of the ring
\begin{equation}
    \mathbf{p} = \frac{\nu\left(\mathbf{v}_{M1}+\mathbf{v}_{M2}\right)}{\left\lVert\mathbf{v}_{M2}-\mathbf{v}_{M1}\right\lVert},\ \text{with}\ \mathbf{v}_{\bullet} = \frac{\mathbf{K}^{-1}[\begin{smallmatrix}\mathbf{x}_{\bullet}\\1\end{smallmatrix}]}{\left\lVert\mathbf{K}^{-1}[\begin{smallmatrix}\mathbf{x}_{\bullet}\\1\end{smallmatrix}]\right\rVert},
\end{equation}
$\mathbf{K}$ the camera intrinsics matrix, and $\nu$ the actual ring radius.
To estimate the normal vector of the \ac{port}, we first need to find the 3D position $\hat{\mathbf{x}}_{m1}$ and $\hat{\mathbf{x}}_{m2}$ of the ellipse's minor axis extremities.
Defining $\hat{\mathbf{x}}_{m\bullet} = d_{m\bullet}\mathbf{v}_{m\bullet}$, with $d_{m\bullet}$ the distance between the camera and $\hat{\mathbf{x}}_{m\bullet}$, allows for the computation of $\hat{\mathbf{x}}_{m\bullet}$ by solving the quadratic problem $\nu^2~=~\left\lVert d_{m\bullet}\mathbf{v}_{m\bullet} - \mathbf{p}\right\rVert^2$.
Note that there are two solutions for each of the extremities of the minor axis, one being closer to the camera than $\mathbf{p}$ and one further.
However, in practice, the ring configuration using the furthest solution with $\mathbf{x}_{m1}$ or the closest solution with $\mathbf{x}_{m2}$ are very similar.
Thus, we only keep the solutions with $\hat{\mathbf{x}}_{m\bullet}$ between the camera position and the ring centre $\mathbf{p}$.
Eventually, these correspond to two different normal vector estimates defined as the cross product between the ellipse's major and minor axis in 3D.

\subsection{Yaw estimation}

Given the aforementioned 5-\ac{dof} estimates, the reflector-filtered image $\mathcal{I}_R$, and a simple \ac{cad} model of the \ac{port}, we remove the ``two-normal" ambiguity and estimate the last \ac{dof} of the \ac{port}'s pose by maximising a correlation score between $\mathcal{I}_r$ and the hypothetical projection of the \ac{port} in the camera frame.
Formally, assuming an orientation $\mathbf{R}$ (rotation matrix) and previously estimated position $\mathbf{p}$, a mask-like image $\mathbf{I}_M = \pi(\mathbf{R},\mathbf{p})$ is created by projecting the \ac{cad}'s reflective markers on the image plane.
Maximising the similarity between $\mathcal{I}_R$ and $\mathcal{I}_M$
\begin{equation}
    \mathbf{R}^* = \underset{\mathbf{R}}{\operatorname{argmin}} \  \sum_{u,v} \pi(\mathbf{R},\mathbf{p})_{[u,v]}\mathcal{I}_{R_{[u,v]}},
    \label{eq:yaw_maximisation}
\end{equation}
provides the final orientation estimate.
Thanks to the prior normal vector estimate, the rotation-based maximisation problem is reduced to two 1-\ac{dof} problems, where the only unknown is the amount of rotation around the normal vectors.
The low dimension of the search space and the \ac{port} axial symmetry allow for the use of a simple grid search between 0 and 120$^\circ$ (increments of 1$^\circ$ in our implementation) to solve~\eqref{eq:yaw_maximisation}.

\section{Experimental setup and data generation}

\subsection{Test bench setup}

\begin{figure}
\def\scale{0.55}
\def\vdist{0.2cm}
\def\hdist{0.2cm}
\def\legendheight{0.77cm}
\def\legendstyle{\footnotesize}
\def\legenddist{0.2cm}
\begin{tikzpicture}
    \tikzstyle{legend} = [fill=white, rectangle, minimum height = \legendheight, align = left, execute at begin node=\setlength{\baselineskip}{8pt}, inner sep=0cm, outer sep=0cm, text width = (1-\scale)*\columnwidth-\hdist,  minimum width = (1-\scale)*\columnwidth-\hdist]
    \tikzstyle{label} = [circle, minimum width=0.5cm, inner sep=0.0cm, outer sep=0cm, fill=white]
    \tikzstyle{labellegend} = [circle, draw=black, minimum width=0.5cm, inner sep=0.0cm, outer sep=0cm, fill=white]

    \node[inner sep=0cm, outer sep=0cm] (img){\includegraphics[width=\scale\columnwidth]{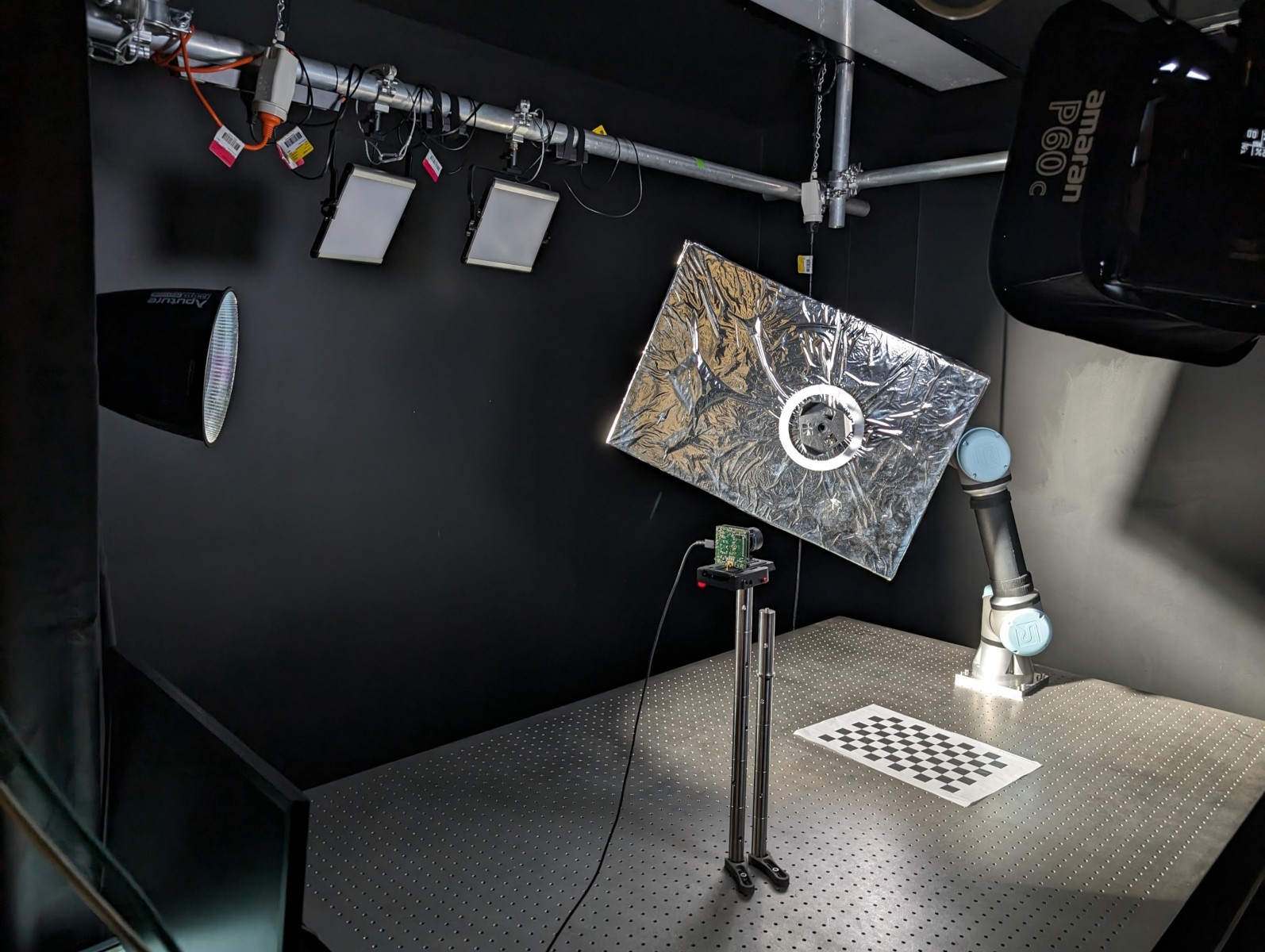}};
    \node[labellegend, right=\hdist of img] (c){\legendstyle c};
    \node[legend, right=\legenddist of c]{\legendstyle Davis 346\\RGB/event camera};
    \node[labellegend, below=\vdist of c] (d) {\legendstyle d};
    \node[legend, right=\legenddist of d]{\legendstyle UR5e robotic arm};
    \node[labellegend, below=\vdist of d] (e) {\legendstyle e};
    \node[legend, right=\legenddist of e]{\legendstyle Aputure P60C\\earthshine light};
    \node[labellegend, above=\vdist of c] (b) {\legendstyle b};
    \node[legend, right=\legenddist of b]{\legendstyle Satellite mock-up\\with docking port};
    \node[labellegend, above=\vdist of b] (a) {\legendstyle a};
    \node[legend, right=\legenddist of a]{\legendstyle Aputure LS1200D\\sunshine light};
    
    \node[label] at (-\scale*3.3cm,\scale*0.1cm){\color{black}a};
    \node[label] at (-\scale*0.2cm,\scale*1.0cm) {\color{black}b};
    \node[label] at (\scale*0.2cm,-\scale*1.1cm) {\color{black}c};
    \node[label] at (\scale*3.0cm,\scale*0.1cm) {\color{black}d};
    \node[label] at (\scale*3.5cm,\scale*1.8cm)  {\color{black}e};
\end{tikzpicture}
\vspace{-0.75cm}
\caption{Photometrically accurate low earth orbit bench for satellite docking.}
\label{figure:test_bench}
\end{figure}
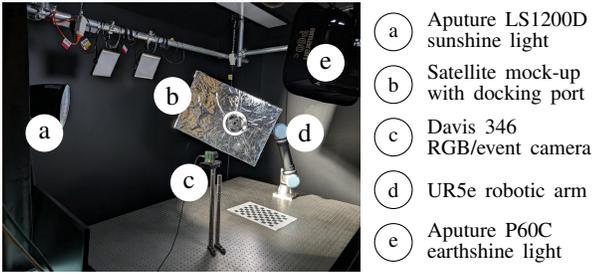

Obtaining data and validating the proposed pipeline on a real satellite in orbit is not practical.
To address this issue, we have developed a photometrically accurate physical simulator~\cite{munasinghe2024towards} as shown in Fig.~\ref{figure:test_bench}.
The setup consists of a satellite mock-up mounted on a robotic arm in a dark room with walls painted in black.
The room is equipped with various light sources to emulate the illumination conditions of \ac{leo}.
One or multiple cameras can be mounted in the environment with an optional colocated light source.
The motion of the robotic arm simulates free-floating relative motion between a servicing satellite (camera) and a target satellite (mock-up with scaled-down \ac{port}).
After performing camera intrinsic and hand-eye calibration, the test bench provides the ground-truth pose of the mock-up in the camera reference frame.
Accordingly, the \ac{cad} model of the \ac{port} can be overlaid onto the camera data to generate training data and to compute quantitative metrics in our experiments.

\subsection{Data generation}

We have recorded around two hours of data with an iniVation Davis 346 across a collection of sequences that last between 40 seconds and 5 minutes and span a wide range of trajectories, mock-up appearance, and illumination conditions.
To conduct various analyses in the following section, we have split the data into four different categories:

\subsubsection{Augmented textures training}
This set contains around one hour and twenty minutes of data with the \ac{port} mounted on non-realistic looking satellite mock-ups (c.f. Fig.~\ref{figure:texture}(a) and~(b)).
The sequences are collected as a combination of \ac{port} trajectories and illumination conditions varying from no light to full sunlighting with earthshine.
\subsubsection{Realistic training}
\begin{figure}
    \centering
    \def\imgwidth{0.32\columnwidth}
    \def\imgdist{0.01\columnwidth}
    \def\subfigdist{0.1cm}
    \begin{tikzpicture}
        \tikzstyle{img} = [fill=white, rectangle, align = center, execute at begin node=\setlength{\baselineskip}{8pt}, inner sep=0, outer sep=0]
        \tikzstyle{subfig} = [fill=white, rectangle, align = center, execute at begin node=\setlength{\baselineskip}{8pt}, inner sep=0, outer sep=0]

        \node[img] (unreal1) {\includegraphics[width=\imgwidth,clip]{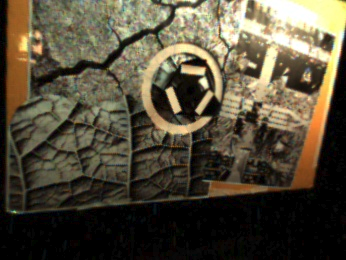}};
        \node[img, right=\imgdist of unreal1] (unreal2) {\includegraphics[width=\imgwidth,clip]{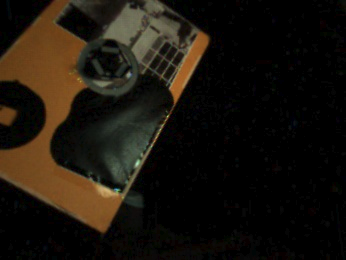}};
        \node[img, right=\imgdist of unreal2] (real) {\includegraphics[width=\imgwidth,clip]{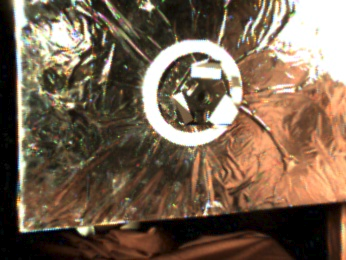}};

        \node[subfig, below=\subfigdist of unreal1]{\subfigstyle (a) Unrealistic cover};
        \node[subfig, below=\subfigdist of unreal2]{\subfigstyle (b) Unrealistic cover};
        \node[subfig, below=\subfigdist of real]{\subfigstyle (c) Realistic mock-up};
    \end{tikzpicture}
    \vspace{-0.25cm}
    \caption{Example of images from the data collected with our satellite docking test bench and the Davis 346 RGB/event camera. (a) and (b) are from the \textit{augmented texture} training set. (c) is from the \textit{realistic training} set.}
    \label{figure:texture}
\end{figure}
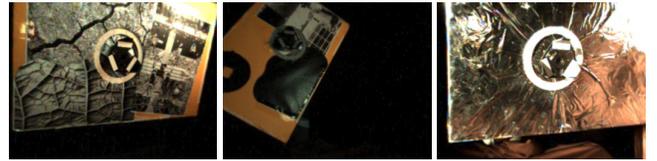

Here, around 20 minutes of data are collected using a realistic satellite mock-up as shown in Fig.~\ref{figure:test_bench} and~\ref{figure:texture}(c).
The mock-up is built with Mylar fixed on an aluminium panel with epoxy to simulate the visual aspect of multi-layer insulation of real satellites.
These data also span different trajectories and illumination conditions.
\subsubsection{Realistic test}
This set consists of three sequences of three and a half minutes using the realistic mock-up.
Each sequence is based on the same trajectory which differs from the ones in \textit{realistic training}.
Only one variable changes from one sequence to another, the level of light: colocated-only, colocated with earthshine, colocated with full sun.
\subsubsection{Hard cases}
Two sequences of 40 seconds have been recorded with the realistic mock-up to show the limits of RGB and event modalities with the proposed pipeline.

\subsection{CNN training}

The proposed \ac{cnn}-based filters are trained in a supervised manner using the known robotic arm pose and satellite mock-up \ac{cad}.
For each RGB image or event histogram, binary masks are independently generated for the navigation ring and the trio of reflectors.
We have trained two sets of ring and reflector filters, one with the \textit{augmented textures} set only, and one with the combination of \textit{augmented textures} and \textit{realistic training} sets.
We respectively denote these \ac{cnn} sets as \textit{high-domain-gap} and \textit{low-domain-gap} models.
The training process simply consists of optimising the \ac{cnn} parameters using the PyTorch implementation of Adam with a binary cross-entropy loss function between the model outputs and the target binary images.
The networks possess around 800k parameters and the training procedure takes less than five hours on an Nvidia RTX A500 laptop GPU using an 80/20 cut for training and validation sets.
Note that we have not performed any specific hyperparameter training on the proposed network.

\section{Experimental results}

In this section, we aim at demonstrating the soundness of the proposed \ac{port} detection and state estimation pipeline as well as comparing the advantages and drawbacks of both RGB and event modalities for the task of satellite docking.
In our quantitative experiments, we filter the outlier state estimates by only considering estimates to be valid when three consecutive predicted poses are close enough to one another.
Concretely, if the orientation differs by more than 15$^\circ$ between consecutive estimates of $\mathbf{R}$, the current pose is rejected.
Results shown in this section include or not this simple filtering step as specified.

All our experiments are run in real-time at 10Hz on a consumer-grade laptop equipped with an Intel i7-1370p CPU with 32GB of RAM, and an Nvidia RTX A500 (mobile) GPU with 4GB of VRAM.
The typical computation load of the proposed pipeline consumes around 25\% of the CPU, 5\% of the GPU, and 330MB of VRAM.
Note that the \ac{ransac} and yaw estimation codes are naive implementations in Python (with the latter code being the main computation bottleneck).
Accordingly, an optimised C++ (with or without GPU parallelisation) would greatly lower the computational cost of the proposed pipeline, allowing for its use on low-power embedded systems.

\subsection{Accuracy and generalisation}

\subsubsection{Gap in domain adaptation}

\begin{table*}[]
    \caption{Accuracy analysis of the proposed detection and state estimation pipeline using RGB or event data in three datasets with different levels of illumination.}
    \centering
    \renewcommand{\arraystretch}{0.5}
\begin{scriptsize}
    \begin{tabular}{lrccccccccccccc}
    & \multicolumn{1}{c}{}& \multicolumn{6}{c}{Using \textit{high-domain-gap} training data} & \multicolumn{1}{c}{} & \multicolumn{6}{c}{Using \textit{low-domain-gap} training data} \\
\cmidrule[\heavyrulewidth]{3-8}
\cmidrule[\heavyrulewidth]{10-15}
& & \multicolumn{2}{c}{Low light} & \multicolumn{2}{c}{Medium light}& \multicolumn{2}{c}{High light}  & $\quad\quad$ & \multicolumn{2}{c}{Low light} & \multicolumn{2}{c}{Medium light}& \multicolumn{2}{c}{High light} \\
& & RGB & Event & RGB & Event & RGB & Event & & RGB & Event & RGB & Event & RGB & Event\\
\midrule
\multicolumn{15}{l}{\textbf{With outlier filtering}}
\\
\midrule
\multirow{2}*{$\quad$Position error [m]} & med. & \textbf{0.011} & 0.013 & \textbf{0.012} & 0.016 & 0.041 & \textbf{0.019} &  & \textbf{0.009} & 0.014 & \textbf{0.011} & 0.014 & 0.019 & \textbf{0.018}\\
& RMSE & \textbf{0.019} & \textbf{0.019} & 0.023 & \textbf{0.022} & 0.054 & \textbf{0.027} &  & \textbf{0.015} & 0.021 & 0.022 & \textbf{0.020} & 0.028 & \textbf{0.024}\\
\cmidrule[\heavyrulewidth]{1-8}
\cmidrule[\heavyrulewidth]{10-15}
\multirow{2}*{$\quad$Normal error [$^\circ$]} & med. & \textbf{4.873} & 5.354 & \textbf{5.616} & 5.656 & 8.084 & \textbf{5.656} &  & \textbf{4.974} & 5.326 & \textbf{4.979} & 5.321 & \textbf{5.276} & 5.682\\
& RMSE & \textbf{5.797} & 6.733 & \textbf{7.761} & 8.061 & 26.066 & \textbf{7.983} &  & \textbf{5.807} & 6.268 & 6.951 & \textbf{6.230} & 7.734 & \textbf{7.257}\\
\cmidrule[\heavyrulewidth]{1-8}
\cmidrule[\heavyrulewidth]{10-15}
\multirow{2}*{$\quad$Rotation error [$^\circ$]} & med. & \textbf{5.719} & 6.603 & \textbf{6.751} & 6.939 & 13.095 & \textbf{7.072} &  & \textbf{5.968} & 6.488 & \textbf{6.063} & 6.742 & \textbf{6.729} & 7.096\\
& RMSE & 9.776 & \textbf{8.498} & \textbf{9.280} & 10.029 & 32.440 & \textbf{9.441} &  & 9.004 & \textbf{7.513} & 8.198 & \textbf{8.005} & 9.588 & \textbf{9.056}\\
\cmidrule[\heavyrulewidth]{1-8}
\cmidrule[\heavyrulewidth]{10-15}
\multirow{2}*{$\quad$Detection rate [\%]} &  all & \textbf{67.4} & 57.1 & 42.0 & \textbf{45.0} & 22.7 & \textbf{39.1} &  & \textbf{69.3} & 38.3 & \textbf{54.3} & 45.9 & \textbf{54.4} & 31.7\\
& in FoV & \textbf{86.9} & 85.5 & 76.5 & \textbf{80.3} & 26.6 & \textbf{66.3} &  & \textbf{89.2} & 65.4 & 81.8 & \textbf{84.1} & \textbf{69.9} & 55.4\\
\midrule
\multicolumn{15}{l}{\textbf{Without outlier filtering}}
\\
\midrule
\multirow{2}*{$\quad$Position error [m]} & med. & \textbf{0.011} & 0.014 & \textbf{0.013} & 0.018 & 0.049 & \textbf{0.021} &  & \textbf{0.010} & 0.015 & \textbf{0.012} & 0.015 & 0.021 & \textbf{0.020}\\
& RMSE & \textbf{0.025} & 0.120 & \textbf{0.031} & 0.098 & 0.084 & \textbf{0.064} &  & \textbf{0.022} & 0.031 & \textbf{0.039} & 0.098 & \textbf{0.039} & 0.083\\
\cmidrule[\heavyrulewidth]{1-8}
\cmidrule[\heavyrulewidth]{10-15}
\multirow{2}*{$\quad$Normal error [$^\circ$]} & med. & \textbf{4.936} & 5.575 & 5.939 & \textbf{5.776} & 8.862 & \textbf{6.154} &  & \textbf{5.021} & 5.538 & \textbf{5.250} & 5.512 & \textbf{5.701} & 6.435\\
& RMSE & \textbf{10.939} & 16.505 & \textbf{10.587} & 19.388 & 31.570 & \textbf{19.168} &  & 10.779 & \textbf{7.851} & 17.401 & \textbf{11.425} & 13.196 & \textbf{13.132}\\
\cmidrule[\heavyrulewidth]{1-8}
\cmidrule[\heavyrulewidth]{10-15}
\multirow{2}*{$\quad$Rotation error [$^\circ$]} & med. & \textbf{6.075} & 6.994 & \textbf{7.230} & 7.484 & 16.887 & \textbf{8.498} &  & \textbf{6.288} & 6.872 & \textbf{6.360} & 7.165 & \textbf{7.472} & 8.410\\
& RMSE & \textbf{15.608} & 18.974 & \textbf{13.206} & 22.844 & 38.952 & \textbf{23.549} &  & 14.747 & \textbf{10.491} & 19.461 & \textbf{15.542} & \textbf{17.030} & 18.213\\
\cmidrule[\heavyrulewidth]{1-8}
\cmidrule[\heavyrulewidth]{10-15}
\multirow{2}*{$\quad$Detection rate [\%]} &  all & \textbf{82.5} & 71.9 & 56.8 & \textbf{62.2} & 58.3 & \textbf{62.2} &  & \textbf{84.8} & 53.8 & \textbf{69.6} & 61.2 & \textbf{75.1} & 56.0\\
& in FoV & \textbf{100} & 99.6 & 98.1 & \textbf{99.2} & 80.1 & \textbf{92.5} &  & \textbf{100} & 90.2 & \textbf{99.7} & 99.4 & \textbf{94.0} & 87.4\\
\midrule
\multicolumn{15}{l}{\textbf{Dataset properties}}
\\
\midrule
\multirow{3}*{$\quad$Saturation of RoI [\%]} & high & 0.0 & - & 1.7 & - & 12.1 & - &  & 0.0 & - & 1.7 & - & 12.1 & - \\
& low & 55.3 & - & 23.2 & - & 0.1 & - &  & 55.3 & - & 23.2 & - & 0.1 & -\\
& total & 55.3 & - & 24.9 & - & 12.2 & - & & 55.3 & - & 24.9 & - & 12.2 & -\\
\cmidrule[\heavyrulewidth]{1-8}
\cmidrule[\heavyrulewidth]{10-15}
\end{tabular}
\end{scriptsize}
\vspace{-0.35cm}
    \label{tab:accuracy}
\end{table*}

This set-up aims to demonstrate the global accuracy of the proposed pipeline as well as pointing out the difference between the modalities in terms of generalisation.
Accordingly, we have performed a quantitative analysis of the framework's output over the three sequences of the \textit{realistic testing} set, and for both the \textit{high-domain-gap} and \textit{low-domain-gap} models, with and without the aforementioned outlier rejection mechanism.
Looking at the results shown in Table~\ref{tab:accuracy}, one can see that with the \textit{high-domain-gap} models and the outlier rejection, both modalities perform similarly for the low and medium-light sequences.
However, with higher illumination, the event-based estimation significantly outperforms the RGB-based one.
Given such results, it could seem that the difference in the cameras' \ac{hdr} capabilities is the main explanation for this empirical observation.
To test this hypothesis, we have run the same experiment using the \textit{low-domain-gap} models.
As shown on the right of Table~\ref{tab:accuracy}, the RGB and event-based results display similar levels of accuracy over all the testing sequences.\footnote{Additional visualisation at \url{https://youtu.be/SuDh-xhnaVY}.}
This rejects the \ac{hdr} difference hypothesis as the accuracy does not correlate with the sensing modality when using a higher level of light.
Thus, we hypothesise that using event-based histograms offers better generalisation abilities than standard RGB data as they are less sensible to the actual appearance of the mock-up due to their ``edge-detection"-like visual aspect.
More precisely, with the low-light sequences the RGB images in both the \textit{augmented texture} and \textit{realistic testing} sets are quite similar (mostly dark) but when using high illumination, the appearance between the two sets differs greatly (larger domain gap).
Thus the \textit{high-domain-gap} models struggle to generalise for RGB data.
This is not the case for the event-based histograms as both the \textit{high-domain-gap} and \textit{low-domain-gap} models perform similarly.
The domain gap hypothesis is also favoured by the fact that in these experiments the level of RGB saturation around the \ac{port} does not correlate, with the final accuracy.

Another interesting observation is the impact of the simple outlier rejection mechanism.
Regardless of the \ac{cnn} model or modality, the outlier filtering does effectively improve the RMSE while having a much lower impact on the median as expected (meaning that less outliers are present in the output).
An obvious drawback of the outlier filtering is the decrease in detection rate.
Note that we also have tested the proposed pipeline on datasets that do not contain any \ac{port}.
We have observed only 5 false detections over more than 5000 processed frames.

\subsubsection{Visual odometry benchmark}

\begin{figure}

    \centering
    \begin{tikzpicture}
        \def\plotwidth{0.45\columnwidth}
        \def\hdist{1em}
        \def\subfigdist{0.1em}
        \def\legendstyle{\tiny}
        \def\legendhdist{4.0em}
        \def\legendxy{-1.3,1.45}

        \tikzstyle{plot} = [fill=white, rectangle, align = center, execute at begin node=\setlength{\baselineskip}{8pt}, inner sep=0, outer sep=0]
        \tikzstyle{subfig} = [fill=white, rectangle, align = center, execute at begin node=\setlength{\baselineskip}{8pt}, inner sep=0, outer sep=0, text width = \plotwidth]

        \node[plot] (plotA) {\includegraphics[width=\plotwidth]{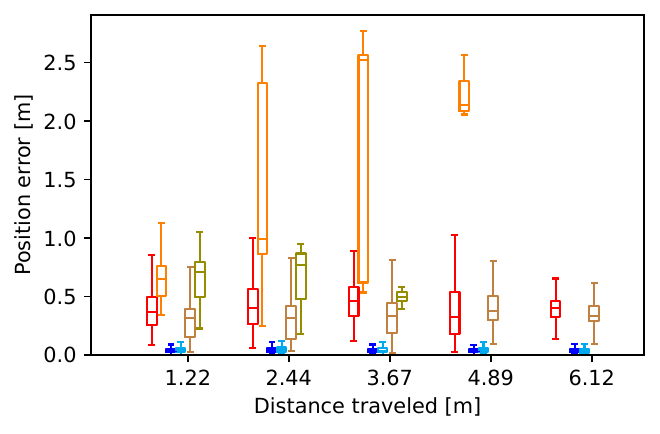}};
        \node[plot, right=\hdist of plotA] (plotB) {\includegraphics[width=\plotwidth]{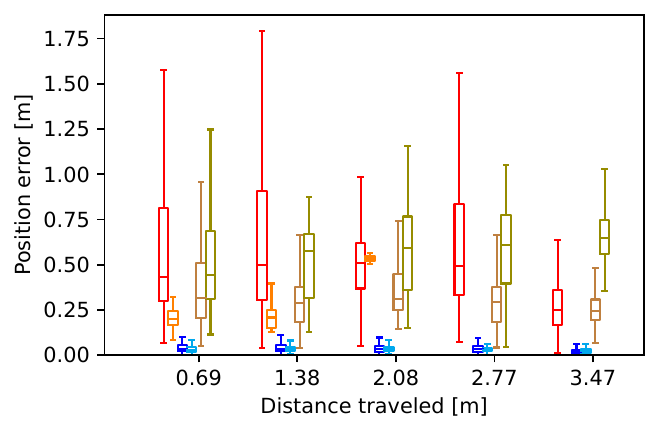}};

        \node[subfig, below=\subfigdist of plotA]{\subfigstyle (a) \textit{Realistic testing} high light};
        \node[subfig, below=\subfigdist of plotB]{\subfigstyle (b) \textit{Realistic training} high light};

        \foreach \i/\c/\n in {0/red/\cite{campos2021orbslam3}-event, 1/orange/\cite{campos2021orbslam3}-RGB, 2/blue/Ours-event, 3/cyan/Ours-RGB, 4/brown/\cite{forster2014svo}-event, 5/olive/\cite{forster2014svo}-RGB}
        {
            \node[draw=\c, minimum width=\legendhdist-0.6em, minimum height=1.1em] at ($(\legendxy)+(\i*\legendhdist,0)$) {\legendstyle\n};
        }
    \end{tikzpicture}
    \caption{Accuracy benchmark with Visual Odometry pipelines over two sequences (a) and (b). The metric is the relative position for various trajectory lengths (based on \cite{zhang2018eval}  with \textit{sim3} alignment of the first 40 frames).} 
    \label{figure:odometry_benchmark}
\end{figure}
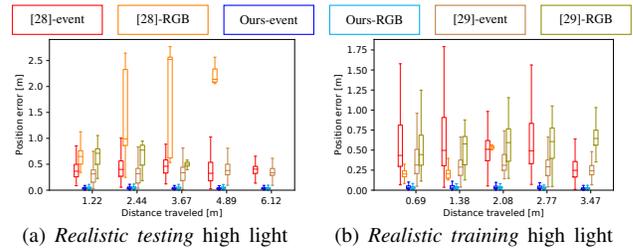

To the best of our knowledge, there is no off-the-shelf open-source estimation pipeline that directly addresses the problem at hand.
To provide a benchmark, we chose to compare our method with common \ac{vo} frameworks that are \textit{EVO}~\cite{rebecq2017evo} (event-only), \textit{OrbSlam3}~\cite{campos2021orbslam3}, and \textit{SVO}~\cite{forster2014svo}.
While our set-up differs from standard \ac{vo}, the background of our test bench is close to being featureless.
This is especially true for the event-based tests as no events are generated by the static background.
Thus, the static-environment assumption of \ac{vo} is not violated and the estimates correspond to the relative pose of the camera with respect to the satellite mock-up.
As \textit{OrbSlam3} and \textit{SVO} are designed to operate with traditional images, we run both frameworks with RGB images and event frames (histograms of 35k consecutive events) independently.
Overall, \textit{EVO} has failed on all the sequences in the \textit{realistic testing} set (tracking failed after a couple of seconds during each test).
\textit{OrbSlam3} and \textit{SVO} best performed on the high-light sequence but also suffered numerous loss of tracking.
As illustrated in Fig.~\ref{figure:odometry_benchmark}, \textit{OrbSlam3} and \textit{SVO} can occasionally perform at a similar level to our method (lowest boundary of the box plots) but the overall accuracy is not sufficient for satellite docking.
For the sake of fairness, it should be noted that the benchmarked methods are not designed for photometrically challenging environments with texture aliasing and rotational symmetries.

\subsubsection{Theoretical limits of model}

\begin{figure}
    \centering
    \begin{tikzpicture}
        \node[] (img) {\includegraphics[width=0.3\columnwidth]{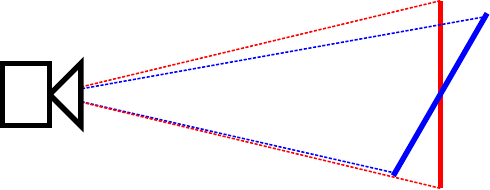}};
        \node[right=-0.3cm of img, yshift=-0.25cm] {\footnotesize\color{red} Ring pose A};
        \node[right=-0.1cm of img, yshift=0.25cm] {\footnotesize\color{blue} Ring pose B};
    \end{tikzpicture}
    \caption{Illustration of the worst case scenario of measurement sensitivity with respect to the docking port inclination.}
    \vspace{-0.4cm}
    \label{figure:theoretical_normal}
\end{figure}

In Table~\ref{tab:accuracy} we have demonstrated the soundness of the proposed estimation pipeline with various metrics.
However, the normal/rotation accuracy does not seem to be on par with the small position error obtained across all datasets with both RBG and event-based modalities.
This observation might be surprising due to the well-known fact that \ac{vo} pipelines generally estimate accurately the camera's orientation with sub-degree precision thanks to the high sensitivity of the camera measurements with respect to the camera orientation. 
Unfortunately, for \ac{port} state estimation, we face the opposite situation where the measurements' sensitivity is very low.
Depending on the relative pose of the port with respect to the camera, a large difference in the \ac{port}'s orientation might lead to a very small variation of measurements.
As illustrated in Fig.~\ref{figure:theoretical_normal} for the worst case scenario (\ac{port} being fronto-parallel to the camera), a one-pixel difference in the \ac{port} appearance at a typical distance of $0.6\,\mathrm{m}$ can be explained by an $8.9^\circ$ difference of orientation.
When the \ac{port} is inclined by $45^\circ$ the same pixel noise corresponds to lower a orientation variation of about $1^\circ$.
Accordingly, the proposed pipeline possesses a lower bound on its rotational accuracy which is a function of the distance to the camera and the true orientation.
This explains why in Table~\ref{tab:accuracy} the position RMSEs are still correct without outlier filtering while the normal and rotation ones are not.

\subsection{Modality limits}

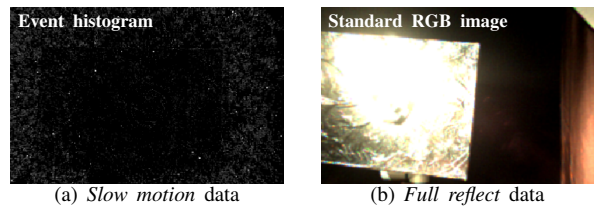
\begin{figure}
    \centering
    \def\imgwidth{0.42\columnwidth}
    \def\imgdist{0.05\columnwidth}
    \def\subfigdist{0.04cm}
    \def\legendstyle{\scriptsize \color{white}}
    \def\legenddist{-0.3cm}
    \def\vdist{0.5cm}
    \begin{tikzpicture}
        \tikzstyle{img} = [fill=white, rectangle, align = center, execute at begin node=\setlength{\baselineskip}{8pt}, inner sep=0, outer sep=0]
        \tikzstyle{subfig} = [rectangle, align = center, text width = \imgwidth,  minimum width = \imgwidth, execute at begin node=\setlength{\baselineskip}{8pt}, inner sep=0, outer sep=0]
        \tikzstyle{legend} = [rectangle, align = left, text width = \imgwidth-0.2cm,  minimum width = \imgwidth-0.2cm, execute at begin node=\setlength{\baselineskip}{8pt}, inner sep=0, outer sep=0]

        \node[img] (eventslow) {\includegraphics[width=\imgwidth,clip, trim=0cm 1.5cm 0cm 0cm]{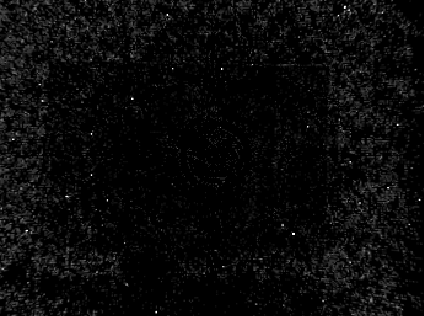}};

        \node[subfig, below=\subfigdist of eventslow.south] {\subfigstyle (a) \textit{Slow motion} data};

        \node[legend, above=\legenddist of eventslow] {\legendstyle\textbf{Event histogram}};

        \node[img, right=\vdist of eventslow] (rgbreflect) {\includegraphics[width=\imgwidth,clip, trim=0cm 1.5cm 0cm 0cm]{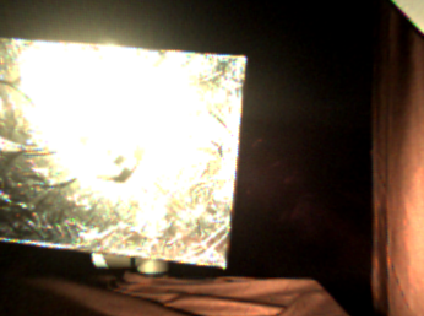}};

        \node[subfig, below=\subfigdist of rgbreflect.south] (subb){\subfigstyle (b) \textit{Full reflect} data};
        
        \node[legend, above=\legenddist of rgbreflect] {\legendstyle\textbf{Standard RGB image}};

    \end{tikzpicture}
    \vspace{-0.25cm}
    \caption{RGB and event data samples from the two  \textit{hard cases} sequences.}
    \label{figure:hardcases}
\end{figure}

With this setup, we briefly expose some of the limitations of the RGB and event modalities.
In Fig.~\ref{figure:hardcases} we show samples from the two sequences, \textit{slow motion} and \textit{full reflect}, from the \textit{hard cases} dataset.
The first one was recorded with extremely low velocity of the target ($1.5\,\mathrm{mm/s}$) and solely the colocated light switched on (no ambient light), while the second one focuses on a high level of reflection from the sun-analog light source on the area of the \ac{port}.
With the \textit{low-domain-gap} models, we obtained a success rate of ellipse detection of 100\% with the RGB data and 1.16\% with the event-based histograms when using the \textit{slow motion} sequence.
For the \textit{full reflect} one, it is the opposite with 0.84\% with the RGB data and 79.8\% with the events.
As illustrated in Fig.~\ref{figure:hardcases}, the RGB camera is incapacitated in the presence of strong reflection due to sensor saturation, and the event frames suffer from an extremely low signal-to-noise ratio when the target moves very slowly.
Both situations are common in the context of satellite docking and maintenance operations.
This emphasises the fact that any single modality is not perfectly fit for our application and that complementing events with RGB data represents a promising path of more robust perception in space applications.


\section{Conclusions}\label{section:conclusion}

We have presented a docking port detection and monocular state estimation framework that mixes data-driven techniques (feature filtering) and geometric models (state estimation).
Unlike most existing methods we focused on satellite-agnostic operations that do not require prior knowledge of the spacecraft's \ac{cad} model, thus leading to better generalisation abilities without the need to retrain any network.
The performance of the framework was evaluated with real data from a physical simulator using both RGB and event cameras.
The proposed method despite possessing theoretical limits demonstrated acceptable levels of accuracy.
However, a conclusion from our experiments is that none of the two modalities is individually sufficient to ensure robust estimation across all the environmental conditions encountered during satellite docking operations.
Accordingly, future work will explore the seamless integration of the RGB and event data to leverage their complementary strengths.
We will also investigate the use of spiking neural networks to better exploit the spatiotemporal nature of event data.

\bibliographystyle{IEEEtran}

\bibliography{references,more_ref}

\end{document}